\documentclass[letterpaper, 10 pt, conference]{ieeeconf}  

\IEEEoverridecommandlockouts                              

\usepackage{amsmath} 
\usepackage{amssymb}  
\usepackage{amsfonts}
\usepackage{soul}
\usepackage{dsfont}
\usepackage[dvipsnames]{xcolor}
\usepackage{graphicx}
\usepackage{tikz}
\usepackage{pgfplots, pgfplotstable}
\usepgfplotslibrary{fillbetween,groupplots}
\usetikzlibrary{shapes,arrows,calc,spy}
\pgfplotsset{grid style={dashed}}

\definecolor{camel}{rgb}{0.76, 0.6, 0.42}%

\definecolor{qual1}{HTML}{7FC97F} 
\definecolor{qual2}{HTML}{BEAED4} 
\definecolor{qual3}{HTML}{FDC086} 
\definecolor{qual4}{HTML}{386CB0} 
\definecolor{qual5}{HTML}{E31A1C} 

\newcommand{\ie}{\textit{i.e.~}} 
\newcommand{\eg}{\textit{e.g.~}} 

\let\mathbb=\mathds 
\newcommand{\vect}[1]{\boldsymbol{#1}}
\newcommand{\E}{\ensuremath{\mathbb{E}}}
\DeclareSymbolFont{matha}{OML}{txmi}{b}{it}
\DeclareMathSymbol{\varv}{\mathord}{matha}{118}

\newtheorem{remark}{Remark}

\newenvironment{list4a}{
	\begin{list}{$\bullet$}{%
			\setlength{\itemsep}{0.05cm}
			\setlength{\labelsep}{0.2cm}
			\setlength{\labelwidth}{0.3cm}
			\setlength{\parsep}{0in} 
			\setlength{\parskip}{0in}
			\setlength{\topsep}{0in} 
			\setlength{\partopsep}{0in}
			\setlength{\leftmargin}{0.16in}}}
	{\end{list}}

\title{\fontsize{17}{19}\selectfont \bf
Maximum Correntropy Criterion Kalman Filter for Indoor Quadrotor Navigation under Intermittent Measurements
}

\author{Loizos Hadjiloizou, Evagoras Makridis, Themistoklis Charalambous, and Kyriakos M. Deliparaschos
\thanks{L. Hadjiloizou is with the Department of Intelligent Systems, KTH Royal Institute of Technology, Stockholm, Sweden
        {\tt\small loizosh@kth.se}.}%
\thanks{E. Makridis and T. Charalambous are with the Department of Electrical and Computer Engineering, School of Engineering, University of Cyprus, Nicosia, Cyprus {\tt\small surname.name@ucy.ac.cy}. T. Charalambous is also with the Department of Electrical Engineering and Automation, School of Electrical Engineering, Aalto University, Espoo, Finland {\tt\small themistoklis.charalambous@aalto.fi}.}%
\thanks{K. M. Deliparaschos is with the Electrical and Computer Engineering and Informatics Department, Cyprus University of Technology, Limassol, Cyprus {\tt\small k.deliparaschos@cut.ac.cy}.}%
}

\begin{document}

\maketitle
\thispagestyle{empty}
\pagestyle{empty}

\begin{abstract}
We present a multisensor fusion framework for the onboard real-time navigation of a quadrotor in an indoor environment. The framework integrates sensor readings from an Inertial Measurement Unit (IMU), a camera-based object detection algorithm, and an Ultra-WideBand (UWB) localisation system. Often the sensor readings are not always readily available, leading to inaccurate pose estimation and hence poor navigation performance. To effectively handle and fuse sensor readings, and accurately estimate the pose of the quadrotor for tracking a predefined trajectory, we design a Maximum Correntropy Criterion Kalman Filter (MCC-KF) that can manage intermittent observations. The MCC-KF is designed to improve the performance of the estimation process when is done with a Kalman Filter (KF), since KFs are likely to degrade dramatically in practical scenarios in which noise is non-Gaussian (especially when the noise is heavy-tailed). To evaluate the performance of the MCC-KF, we compare it with a previously designed Kalman filter by the authors. Through this comparison, we aim to demonstrate the effectiveness of the MCC-KF in handling indoor navigation missions. The simulation results show that our presented framework offers low positioning errors, while effectively handling intermittent sensor measurements.
\end{abstract}

\section{INTRODUCTION}
The recent advancements in wireless communications and computer vision for multi-rotor Unmanned Aerial Vehicles (UAVs) have enabled fully autonomous navigation in both outdoor and indoor environments. Time- and safety-critical missions such as search and rescue, facility monitoring, and warehouse inventory management require reliable information about the UAV's pose (\ie position and orientation), which includes its position and orientation. To achieve this, the pose of the quadrotor needs to be estimated reliably and with high precision, especially for indoor missions where accuracy, robustness, and timely reactions to changes in the environment are crucial.

While there are advanced techniques for multisensor data fusion, they typically combine cues from Inertial Measurement Unit (IMU) devices and cameras for Simultaneous localisation and Mapping (SLAM). However, these visual-inertial approaches tend to accumulate errors in the pose over time due to noise in the sensors and modeling errors. This problem can be addressed in outdoor environments by incorporating global position measurements such as GPS, as seen in studies like \cite{mascaro2018gomsf}, while in indoor environments, it is more practical to have pre-existing mapping and features with known locations to improve the localisation.

Vision-based approaches for localising UAVs use both off-board and on-board visual sensing methods. Off-board sensing typically relies on expensive fixed motion-capture systems with high-frame rate cameras, as seen in studies like \cite{preiss2017crazyswarm,sa2018dynamic}, making it less portable and more cumbersome to use. On-board sensing, on the other hand, uses on-board cameras such as monocular cameras \cite{lim2015monocular,du2020real} and stereo cameras ~\cite{fraundorfer2012vision}, which offer a more practical and cost-effective solution. However, these on-board methods have limitations in performance under changes in illumination and viewpoint \cite{maffra2019real}. UAV localisation using wireless technology, specifically Ultra-Wideband (UWB), has been a topic of interest among researchers and practitioners due to its scalability, cost-effectiveness, and ease of installation \cite{papastratis2018indoor, makridis2020towards}. Despite its advantages, UWB-based localisation can be unreliable in certain conditions, such as in the absence of line-of-sight, which can result in noisy measurements and communication delays. Multiple techniques for integrating data from UWB localisation systems and SLAM have been proposed to achieve a more accurate and drift-free estimate of a robot's position in indoor environments \cite{tiemann2018enhanced, yang2021uvip, nguyen2021range}. In particular, the authors in \cite{yang2021uvip} were pioneers in proposing a multisensor fusion approach that can accommodate sensor failures. Their proposed positioning system is based on optimisation-based sensor fusion, which is independent of the quadrotor state model and can enhance positioning accuracy and robustness, even in the presence of sensor failures.

Although these works consider multisensor approaches with measurements driven by Gaussian noise, in practice, measurements from sensors are often intermittently available and disturbed by impulsive (shot) noises due to sensor's incapability to provide reliably its measurements. In our recent work \cite{hadjiloizou2022onboard}, we designed a multisensor pose estimation method based on Kalman filter to handle such intermittent observations that are driven by Gaussian noise. Moreover, the authors in \cite{CHEN201770} employed a Kalman filter based on the maximum correntropy criterion which is capable of handling non-Gaussian noise. Other works utilising the maximum correntropy criterion for state estimation include~\cite{zhou2022adaptive} and~\cite{Zhao:2022}, where the authors proposed a multisensor fusion algorithm based on the Unscented Kalman filter combining cues from an IMU and UWB.

In this work, we propose a quadrotor pose estimation framework that handles measurements from multiple sensors that arrive intermittently at the estimator, while they are driven by non-Gaussian noise. In particular, in this work we make the following contributions.
\begin{list4a}
\item We deploy a Maximum Correntropy Criterion Kalman Filter (MCC-KF) for fusing measurements from an IMU, a camera-based object detection algorithm, and an UWB localisation system. The MCC-KF is modified in two different ways to handle intermittent sensor readings in order to provide robust pose estimation to a Linear Quadratic Servo controller such that a predefined trajectory is followed, albeit the absence or abrupt variation of the measurements. Specifically, when measurements are not received, the maximum correntropy criterion formula, which involves the measurements obtained, either uses the previous measurement or the innovation of the estimate is assumed zero. 
\item It is shown numerically that either of the two approaches proposed herein perform better than the classical KF with intermittent measurement (see Section~\ref{subsec:results}). Nevertheless, the one that uses the previous measurement as the current one obtains better results than the one in which the innovation is set to zero.
\end{list4a}

\textbf{Notation.} In this paper, bold lowercase letters are used to denote vectors, uppercase letters for matrices, and calligraphic uppercase letters for sets. The sets $\mathbb{R}$, $\mathbb{R}{+}$, and $\mathbb{N}$ represent real, non-negative real, and natural numbers, respectively. The identity matrix of dimension $p$ is represented as $I{p\times p}$ or simply as $I$ if its dimensions are clear. The transpose of matrix $A\in \mathbb{R}^{p\times m}$ is denoted as $A^{T}$, and its inverse as $A^{-1}$ if $m=p$. The diagonal elements of a matrix are represented by $\mathrm{diag}{A}$. A positive semi-definite matrix is denoted as $A\succeq 0$ and a positive definite matrix as $A\succ 0$. The expected value of a quantity is denoted as $\mathbb{E}{\cdot }$. The sine and cosine of an angle $\theta$ are denoted as $s\theta$ and $c\theta$, respectively.

\section{SYSTEM DESCRIPTION}
\subsection{Coordinate Systems}
To determine the location of the quadrotor in space and the relative locations of surrounding objects around it, we define two coordinate frames using the standard right-handed robotics convention as shown in Fig.\ref{fig:quadrotor}. The Earth's inertial frame $\{E\}$ follows the East-North-Up (ENU) reference system where $+x$ axis points to the east, $+y$ to the north and $+z$ points upwards based on the right-hand rule. The Body frame of the quadrotor $\{B\}$ is coincident to the origin and thus to the absolute position of the quadrotor, \ie $[x,~y,~z]$, while it follows the Forward-Left-Up (FLU) which gives forward horizontal, left horizontal and up vertical movement along its $+x$, $+y$ and $+z$ axis respectively.

\subsection{Quadrotor Model}
\begin{figure}[t]
	\begin{center}
            \vspace{0.11cm}
		\includegraphics[width=0.95\columnwidth]{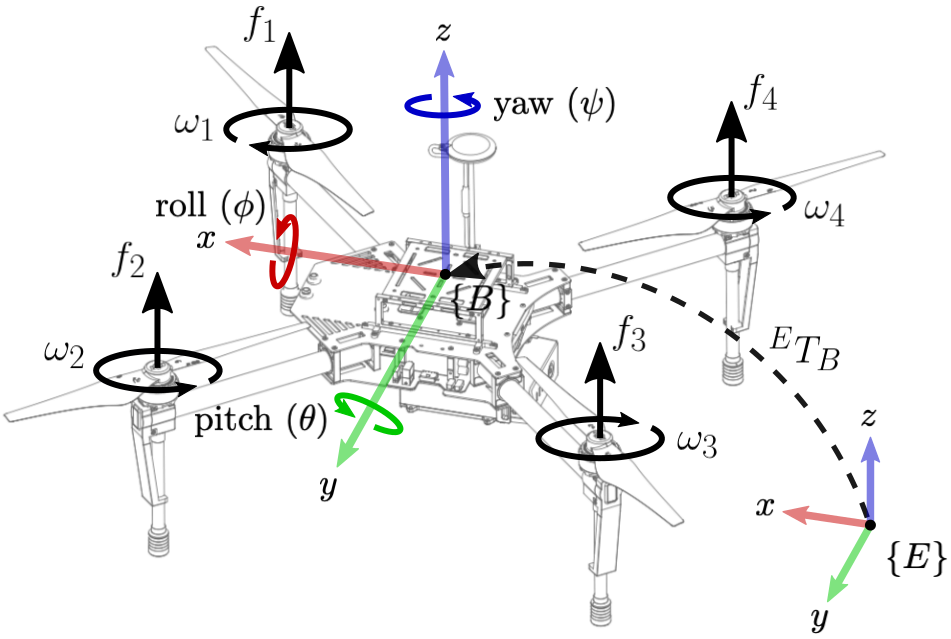}
		\caption{DJI M100 quadrotor model with coordinate frame.}
		\label{fig:quadrotor}
	\end{center}
\end{figure}

The quadrotor is a naturally unstable non-linear complex system composed of four rotors, as the name suggests. Each rotor is made up of a propeller and a motor that generates an angular velocity $\omega_i$, resulting in a thrust force $f_i$, where $i$ refers to the number of the motor as depicted in Fig.\ref{fig:quadrotor}. Two rotors rotate clockwise while the other two rotate counterclockwise to prevent unwanted rotation of the quadrotor body in the yaw ($\psi$) direction (conservation of angular momentum). The angular velocities of the rotors correspond to specific rotational coordinates, $\vect{\eta}=(\phi,\theta,\psi)\in\mathbb{R}^3$, and move the quadrotor to different translational coordinates, $\vect{\xi}=(x,y,z)\in\mathbb{R}^3$, in the Earth inertial frame ${E}$. The orientation of the quadrotor is defined by the Euler angles $\phi$, $\theta$ and $\psi$. $\phi$ (roll) is the angle around the $x$-axis, $\theta$ (pitch) is the angle around the $y$-axis, and $\psi$ (yaw) is the angle around the $z$-axis. The translational coordinates $x$, $y$ and $z$ represent the center of mass of the quadrotor relative to the Earth inertial frame.

The translational and rotational equations of motion for a quadrotor in Earth frame are described using the Newton-Euler formalism \cite{kendoul2006nonlinear,voos2009nonlinear} and \cite{sabatino2015quadrotor}, respectively. These equations take into account relatively small quadrotor movement angles
\begin{subequations}\label{eq:eom}
\begin{align}
    \ddot{x} &= \frac{f_T}{m} \left(c\phi s\theta c\psi + s\psi s\phi \right), \quad     \ddot{\phi}=\frac{I_{y}-I_{z}}{I_{x}} \dot{\theta} \dot{\psi}+\frac{\tau_{x}}{I_{x}},\nonumber\\
	\ddot{y} &= \frac{f_T}{m} \left(c\phi s\theta s\psi - c\psi s\phi \right),\quad     \ddot{\theta}=\frac{I_{z}-I_{x}}{I_{y}} \dot{\phi} \dot{\psi}+\frac{\tau_{y}}{I_{y}},\nonumber\\
	\ddot{z} &= \frac{f_T}{m} \left(c\phi c\theta \right) - g, \quad\;\;\; \quad\quad    \ddot{\psi}=\frac{I_{x}-I_{y}}{I_{z}} \dot{\phi} \dot{\theta}+\frac{\tau_{z}}{I_{z}},\nonumber
\end{align}
\end{subequations}
where $m$ represents the mass of the quadrotor, $f_T$ represents the total thrust force, and $g$ represents the acceleration due to gravity. The terms $\tau_\theta$, $\tau_\phi$, and $\tau_\psi$ are the pitch torque, roll torque, and yaw torque respectively and they depend on the angular velocities of the rotors $\Omega_i$ \cite{bouabdallah2007design}. Furthermore, $I_x$, $I_y$, and $I_z$ are the moments of inertia of the quadrotor's symmetric rigid body around its three axes.

A simplified linearised model can be obtained around the hovering equilibrium point by small-angle approximation (\ie assuming that the rotational angles of the system are relatively small), which implies that $\dot{\phi},\dot{\theta},\dot{\psi}\simeq 0$, and $s\phi \simeq \phi,s\theta \simeq \theta,s\psi \simeq \psi$, with $c\phi = 1$, $c\theta = 1$, and $c\psi = 1$. The system state vector is defined as $\small \vect{x}=[x~y~z~\phi~\theta~\psi~\dot{x}~\dot{y}~\dot{z}~\dot{\phi}~\dot{\theta}~\dot{\psi}]^T$. The hovering equilibrium point, $\vect{\bar{x}}=[\bar{x}~\bar{y}~\bar{z}~0~0~0~0~0~0~0~0~0]^T$, is reached when the total thrust is a constant control input of $f_T=mg$, reflecting the force necessary to hover the quadrotor at an arbitrary position $(\bar{x},\bar{y},\bar{z})$. The resulting linearised system is described by
\begin{align}
\ddot{x}&=g \theta,\quad \ddot{y}=-g \phi, \quad \ddot{z}= \frac{f_{T}}{m},\nonumber\\
\ddot{\phi}&=\frac{\tau_{x}}{I_{x}},\;\;\; \ddot{\theta}=\frac{\tau_{y}}{I_{y}}, \quad\;\; \ddot{\psi}=\frac{\tau_{z}}{I_{z}}.
\end{align}

\subsection{IMU}
An IMU, or Inertial Measurement Unit, is a device that uses a combination of accelerometer, gyroscope, and magnetometer sensors to determine a body's orientation, velocity, and gravitational forces. It is often used in conjunction with other sensors, such as those based on vision or wireless technology, to enhance the accuracy of pose estimation. In some cases, the IMU may serve as the primary sensor if other sensors are not available. The IMU provides measurements of a quadrotor's orientation using the three Euler angles: roll, pitch, and yaw.

\subsection{UWB Localisation}
Ultra-Wideband (UWB) is a wireless communication technology that utilises short pulses with low energy over a wide bandwidth, making it highly resistant to multipath interference. Its unique characteristics also enable precise measurement of Time-of-Flight (ToF), which allows for accurate distance estimation. These features make UWB a popular technology for localisation in indoor environments. localisation using UWB involves placing wireless transmitters, known as anchors, at specific locations and using a UWB receiver, known as a tag, on the quadrotor to log the arrival times of UWB signals and calculate the quadrotor's position in space \footnote{The locations of the anchors can also be inferred if they are in a certain formation and the initial position of the tag is known.}.

\subsection{Camera-Based Localisation}
Camera-based detection of landmarks is a well-established approach for robot localisation. The method involves utilising a camera mounted on the quadrotor and a set of landmarks whose positions are known with respect to the global reference frame. Analysing the objects detected by the camera facilitates the estimation of the quadrotor's position. Given that a monocular camera is used, depth information about the landmarks can be obtained by analysing successive frames. This process enables the estimation of the relative position between a landmark and the camera. Once the relative position of the camera is known, the relative position of the quadrotor on which the camera is attached to, can be obtained and expressed in the global reference frame.

\section{POSE ESTIMATION AND CONTROL}
The continuous-time state-space representation of the quadrotor, is defined as follows
\begin{equation}
\begin{aligned}
\mathrm{d}\vect{{x}}(t) &= A \vect{x}(t)\mathrm{d}t + B \vect{u}(t) \mathrm{d}t + \mathrm{d}\vect{w}(t),\\
\mathrm{d}\vect{y}(t) &= C\vect{x}(t)\mathrm{d}t+\mathrm{d}\vect{v}(t),
\end{aligned}
\end{equation}
where $\vect{x} \in\mathbb{R}^{12}$ is the system state vector, \ie $\vect{x}=[x~y~z~\phi~\theta~\psi~\dot{x}~\dot{y}~\dot{z}~\dot{\phi}~\dot{\theta}~\dot{\psi}]^{\top}$, $\vect{u} \in\mathbb{R}^{4}$ is the control input vector, i.e., $\vect{u}=[f_{T}~\tau_{x}~\tau_{y}~\tau_{z}]^{\top}$, $\vect{y} \in\mathbb{R}^{9}$ is the measurement vector, i.e., $\vect{y}=[x_{\scriptscriptstyle uwb}~y_{\scriptscriptstyle uwb}~z_{\scriptscriptstyle uwb}~x_{\scriptscriptstyle cam}~y_{\scriptscriptstyle cam}~z_{\scriptscriptstyle cam}~\phi~\theta~\psi]^{\top}$, while $\vect{w}\in\mathbb{R}^{12}$ and $\vect{v}\in\mathbb{R}^{9}$ are zero-mean stochastic processes, representing the process and measurement noise levels. The discrete-time equivalent state-space model of the quadrotor, can be expressed as
\begin{equation}\label{eq:lti_system}
\begin{aligned}
\vect{x}_{k+1} &= \Phi \vect{x}_k + \Gamma \vect{u}_k + \vect{w}_k,\\
\vect{y}_k &= C\vect{x}_k + \vect{v}_k,
\end{aligned}
\end{equation}
\noindent where the discrete-time equivalent system matrix is represented by $\Phi=e^{Ah}$, the control input matrix is given by $\Gamma=\int_{s=0}^{h} e^{As} B ds$, and $h$ is the sampling period, also known as the control loop interval. The measurement matrix $C \in \mathbb{R}^{p\times n}$ defines the connection between the system states and the readings taken by the sensors, which make up the system output. The process noise and measurement noise are represented by $\vect{w}_k$ and $\vect{v}_k$, respectively, and are assumed to be zero-mean white Gaussian random sequences, with $\E\{\vect{w}_k\}=0$, $\E\{\vect{v}_k\}=0$, $\E\{\vect{w} \vect{w}^T\} = W\succeq0$, and $\E\{\vect{v} \vect{v}^T\} = V\succ0$. We represent the state estimates as $\hat{\vect{x}}_{k|k-1}$ (a priori) and $\hat{\vect{x}}_{k|k}$ (a posteriori), with the error covariance matrices defined by
\begin{align}
P_{k|k-1} &\triangleq \E \{(\vect{x}_k - \hat{\vect{x}}_{k|k-1})(\vect{x}_k - \hat{\vect{x}}_{k|k-1})^T\} ,\\
P_{k|k} &\triangleq \E \{(\vect{x}_k - \hat{\vect{x}}_{k|k})(\vect{x}_k - \hat{\vect{x}}_{k|k})^T\}.
\end{align}

\subsection{Linear Quadratic Servo (LQ-Servo) Control}\label{sec:lqr}
To follow a reference signal $\vect{r}_k = [r_x,r_y,r_z]^T$ that is generated by the path planner, we augment the state-space model in \eqref{eq:lti_system} by adding an integral state vector
$$
\vect{i}_{k+1} = \vect{i}_k + \vect{r}_k - E \vect{y}_{k}=-EC \vect{x}_k + \vect{i}_k + \vect{r}_k - E\vect{v}_k,
$$
where $E \in \mathbb{R}^{3 \times {9}}$ is the matrix that selects the controlled states (\ie $E = [\Delta_{1, k}, ( \sim \Delta_{1, k} \wedge (\Delta_{2, k}) ), 0_{3\times3}]$), with $(\sim)$ and $(\wedge)$ being the element-wise logical NOT and AND operators, respectively. Thus, the augmented state-space model is
{\small 
\begin{align}
 \underbrace{\begin{bmatrix} \vect{x}_{k+1} \\ \vect{i}_{k+1} \end{bmatrix}}_\text{{\small $\vect{\bar{x}}_{k+1}$}}
 \!&=\!
  \underbrace{
  \begin{bmatrix}
	\Phi & 0 \\
	-EC & I
   \end{bmatrix}}_\text{{\small $\bar{\Phi}$}}\!
   \underbrace{\begin{bmatrix}
	\vect{x}_{k} \\ \vect{i}_{k} 
   \end{bmatrix}}_\text{{\small $\vect{\bar{x}}_{k}$}}\!
    +\!
   \underbrace{\begin{bmatrix}
	\Gamma \\
	0
   \end{bmatrix}}_\text{{\small $\bar{\Gamma}$}}\!
\vect{u}_k +\!
   \underbrace{\begin{bmatrix}\!
	I & 0\\
	0 & -E
   \end{bmatrix}}_\text{{\small $\bar{E}$}}\!
   \underbrace{\begin{bmatrix}\!
	\vect{w}_k \\
	\vect{v}_k
   \end{bmatrix}}_\text{{\small $\bar{\varv}_k$}} +\!
    \underbrace{\begin{bmatrix}
	0 \\ I
   \end{bmatrix}}_\text{{\small $\bar{I}$}}\!
\vect{r}_k,\nonumber\\
\vect{\bar{y}}_k &= \underbrace{\begin{bmatrix}
    C & 0
\end{bmatrix}}_\text{{\small $\bar{C}$}}
\underbrace{\begin{bmatrix}
    \vect{x}_k \\ \vect{i}_k
\end{bmatrix}}_\text{{\small $\bar{\vect{x}}_k$}}\!+\: \vect{v}_k,\nonumber
\end{align}}%
where $\bar{\varv}_k$ and $\vect{v}_k$ are discrete-time Gaussian white noise processes with zero-mean value and {covariances} $\mathbb{E}\left\{\bar{\varv}_k \bar{\varv}_k^{T}\right\}=V_{1}$, $\mathbb{E}\left\{\bar{\varv}_k \vect{v}_k^{T}\right\}=V_{12}$, and $\mathbb{E}\left\{\vect{v}_k \vect{v}_k^{T}\right\}=V_{2}$,
\begin{align}
\mathbb{E}\left\{\left[\begin{array}{l}
    \bar{\varv}_k \\
    \vect{v}_k
    \end{array}\right]\left[\begin{array}{l}
    \bar{\varv}_k \\
    \vect{v}_k
    \end{array}\right]^{T}\right\}=\left[\begin{array}{cc}
    V_{1} & V_{12} \\
    V_{12}^{T} & V_{2}
    \end{array}\right].\nonumber
\end{align}
However, since $\bar{\varv}_k$, and $\vect{v}_k$ are correlated, then 
\begin{align}
    \mathbb{E}\left\{\bar{\varv}_k \vect{v}_k^{T}\right\}=V_{12}=\mathbb{E}\left[\begin{array}{ll}
    \vect{w}_{k} \vect{v}_{k}^{T} & \vect{v}_{k} \vect{v}_{k}^{T}
    \end{array}\right]=\left[0 \quad V_{2}\right].\nonumber
\end{align}
As long as the availability of the measurements are independent of the control inputs, which is the case in this work,  the certainty equivalence principle holds \cite{Molin:2013, Ramesh:2011, Farjam2022}.  Therefore, the optimal control law $\vect{u}_k^* = -L\vect{\bar{x}}_{k} = -L^{\hat{x}} \vect{\hat{x}}_{k|k-1} - L^{i} \vect{i}_k$ can be found by minimising the linear quadratic criterion in \eqref{eq:lq_cost}
\begin{align}
J &= \mathbb{E}\left[\vect{\bar{x}}_{N}^{\mathrm{T}} \bar{Q}_N \vect{\bar{x}}_{N}+\sum_{k=0}^{N-1}\left(\vect{\bar{x}}_{k}^{\mathrm{T}} \bar{Q}_{k} \vect{\bar{x}}_{k}+\vect{u}_{k}^{\mathrm{T}} R_{k} \vect{u}_{k}\right)\right],\label{eq:lq_cost}
\end{align}
where $\bar{Q}_N\succeq 0$, $\bar{Q}_{k}\succeq 0$ are the final and stage state error weighting matrices, respectively, and $R_{k}\succ0$ is the stage control weighting matrix for the LQ problem. The optimal control gain $L_k = \begin{bmatrix} L_k^{\hat{x}} & L_k^{i}\end{bmatrix}$ is the standard state-feedback controller gain given by
\begin{align}
L_{k}=\left(\bar{\Gamma}^{T} S_{k+1} \bar{\Gamma}+R_k\right)^{-1} \bar{\Gamma}^{T} S_{k+1} \bar{\Phi},
\end{align}
and where $S_{k}$ satisfies the  discrete-time algebraic Riccati equation (DARE) 
\begin{align} \label{eq:DARE}
S_{k} &= \bar{Q}_k + \bar{\Phi}^{T} S_{k+1} \bar{\Phi} \nonumber\\
&\quad-\bar{\Phi}^{T} S_{k+1} \bar{\Gamma}\left(\bar{\Gamma}^{T} S_{k+1} \bar{\Gamma} + R_k\right)^{-1} \bar{\Gamma}^{T} S_{k+1} \bar{\Phi} . 
\end{align}
By considering the infinite horizon problem with $\bar{Q}_k = \bar{Q}$ and $R_k = R$ for all time steps $k$, and based on the assumption that the pairs $(\bar{\Phi},\bar{\Gamma})$ and $(\bar{\Phi},\bar{Q}_k^{1/2})$ are controllable and observable, respectively, the positive semi-definite solution of \eqref{eq:DARE} always exists \cite{Chen:1995}. Then, the controller $L_{\infty}$ becomes
\begin{align}
L_{\infty} = \left(\bar{\Gamma}^{T} S_{\infty} \bar{\Gamma}+R\right)^{-1} \bar{\Gamma}^{T} S_{\infty} \bar{\Phi} ,
\end{align}
where $S_{\infty}$ is the positive semi-definite solution of the DARE
\begin{align} \label{eq:DARE-infty}
S_{\infty} &= \bar{Q} + \bar{\Phi}^{T} S_{\infty} \bar{\Phi} -\bar{\Phi}^{T} S_{\infty} \bar{\Gamma}\left(\bar{\Gamma}^{T} S_{\infty} \bar{\Gamma} + R\right)^{-1} \bar{\Gamma}^{T} S_{\infty} \bar{\Phi}.\nonumber
\end{align}

\subsection{MCC Kalman Filter for Multisensor Fusion}
In this section, we present the Maximum Correntropy Criterion Kalman Filter (MCC-KF) and how is used/modified to account intermittent measurements for establishing pose estimation for the quadrotor. MCC-KF is often used to deal with non-Gaussian noises, \eg shot noise or mixture of Gaussian noise \cite{CHEN201770,izanloo2016kalman}, by measuring the similarity of two random variables using information from high-order signal statistics.

The (by now) classical MCC-KF filter determines a set of observer gains $K_k$ based on the maximum correntropy criterion $K^{\text{mcc}}_k$ to minimise the estimation error covariance. For the estimation error, $\tilde{\vect{x}}_{k+1}\triangleq \bar{\vect{x}}_{k+1}-\hat{\vect{x}}_{k+1|k}$, we have
\begin{align}
\tilde{\vect{x}}_{k+1}=\bar{K}_k \left(\left[\begin{array}{l}
\bar{\Phi} \\
H_k
\end{array}\right] \tilde{\vect{x}}_{k}+\left[\begin{array}{ll}
\bar{E} & 0 \\
0 & I
\end{array}\right]\left[\begin{array}{l}
\bar{\varv}_{k} \\
\vect{v}_{k}
\end{array}\right]\right)+\bar{I} \vect{r}_{k}, \nonumber
\end{align}
where $\bar{K}_k \triangleq \left[I \quad -K_k\right]$. The error covariance of the augmented state-space model is
\begin{align}
P_{k+1} &=\mathbb{E}\left\{\tilde{x}_{k+1}{\tilde{x}}_{k+1}{ }^{T}\right\} \nonumber \\
&=\bar{K}_k\left[\begin{array}{ll}
\bar{\Phi} P_{k} \bar{\Phi}^{T}+\bar{E} V_{1} \bar{E}^{T} & \bar{\Phi} P_{k} H_k^{T}+\bar{E} V_{12} \\
H_k P_{k} \bar{\Phi}^{T}+V_{12} \bar{E}^{T} & H_k P_{k} H_k^{T}+V_{2}
\end{array}\right] \bar{K}_k^T. \nonumber
\end{align}
Hence, the MCC-KF filter for the augmented state-space (which fuses all the measurements) becomes
\begin{subequations}
	\begin{flalign}
    K^{\text{mcc}}_k & = \frac{{{G_\sigma }\left( {\parallel {\vect{y}_{k}} - H_k{{\hat{\vect{x}} }_{k|k-1}}{\parallel _{V_2^{ - 1}}}} \right)}}{{{G_\sigma }\left( {\parallel  \hat{\vect{x}} _{k+1|k}  - \bar{\Phi}{{\hat{\vect{x}} }_{k|k} {-\bar{\Gamma}\vect{u}_{k}} }{\parallel _{P_{k+1}^{ - 1}}}} \right)}}, \nonumber \\
    K_{k} &= \left(\bar{\Phi} P_{k} K^{\text{mcc}}_k H_k^{T}+\bar{E} V_{12}\right) \left(H_k P_{k} K^{\text{mcc}}_k H_k^{T}+V_{2}\right)^{-1},\nonumber\\
    P_{k+1} &=\bar{\Phi} P_{k} \bar{\Phi}^{T}+\bar{E} V_{1} \bar{E}^{T} - \left(\Phi P_{k} H_k^{T}+\bar{E} V_{12}\right)\nonumber\\
    &\quad\quad \left(H_k P_{k} H_k^{T}+V_{2}\right)^{-1}\left(H_k P_{K} \bar{\Phi}^{T}+V_{12} \bar{E}^{T}\right),\nonumber\\
    \hat{\vect{x}}_{k+1|k} &= \bar{\Phi} \hat{\vect{x}}_{k|k-1} +\bar{\Gamma} \vect{u}_{k}+K_{k}\left(\bar{\vect{y}}_{k}-{H}_k \hat{\vect{x}}_{k|k-1}\right).\nonumber
	\end{flalign}%
\end{subequations}
where $G_\sigma$ is the Gaussian kernel, \ie
\begin{align*}
	G_{\sigma}( \parallel \vect{x}_i-\vect{y}_i \parallel )=\exp\left(-\frac{\parallel \vect{x}_i-\vect{y}_i \parallel ^2}{2\sigma^2}\right),
\end{align*}
with kernel size $\sigma$ (representing a weighting parameter between the second and higher-order moments). Note that $K^{\text{mcc}}_k$ is the \emph{minimised correntropy estimation cost function} and $K_k$ is the Kalman gain.

While the MCC-KF has received a tremendous attention (see, \eg \cite{Qi_2023,liao_dynamic_2022} and references therein), to the best of the authors knowledge, no work has considered how possible absence of measurements are handled by the filter. In this paper, we consider two candidate approaches, which we evaluate for the application we consider in Section~\ref{subsec:results}:
\begin{list4a}
\item[1.] In the first approach, the measurements not received are replaced by their previous measurements instead of setting them to zero. The rationale for this choice is to guarantee that the Gaussian kernel does not have abrupt variations between consecutive steps, thus preventing large fluctuations of the Kalman gain (which in some cases approaches zero due to the large norm of the expected measurement $H_k{{\hat{\vect{x}} }_{k|k-1}}$). Let $\vect{y}_{k}^j$ denote the measurement vector as observed by sensor $j$ at time instance $k$, where $j$ denotes the index of the sensor (\ie IMU, UWB, and camera). Then, we define the measurement vector $\vect{y}_{k}^j$ as
\begin{align}
    \vect{y}_{k}^j = \delta_{k}^{j} \vect{y}_{k}^{j} + (1 - \delta_{k}^{j}) \vect{y}_{k - 1}^{j}, 
\end{align}
where $\delta_{k}^{j}=1$ if measurement from sensor $j$ is available at time step $k$, and $\delta_{k}^{j}=0$, otherwise.
\item[2.] In the second approach, we use the expected measurement as the actual measurement, \ie
\begin{align}
    \vect{y}_{k}^j = \delta_{k}^{j} \vect{y}_{k}^{j} + (1 - \delta_{k}^{j}) H_k{{\hat{\vect{x}} }_{k|k-1}}.
\end{align}
The rationale for this choice is that the estimate serves as the best guess for the actual value of the missing measurement.
\end{list4a}

\begin{remark}
Note that the controller is pre-specified to be the optimal state-feedback controller for the LQ problem. Since the minimised correntropy estimation cost function $K^{\text{mcc}}_k$ depends on the controller and affects the filter gain, it is expected that the certainty equivalence principle does not hold. 
\end{remark}

%
%

\section{EXPERIMENTAL VALIDATION}
\subsection{Experimental setup}
Our approach in this work involves utilising the Webots open-source 3D robotics simulator tool to model the proposed system. We augment the environment with several UWB tags and objects, detectable by the integrated object detection algorithm of the simulator. The control system alongside the estimators and any auxiliary functions are implemented as Robot Operating System (ROS) nodes in Python\footnote{For reproducibility, we are sharing the core project code via the following GitHub repository https://github.com/loizoshad/MCCKFuIM}. The communication link between the simulated quadrotor, the controller and estimator is based on the ROS message exchange framework.

In our configuration, the quadrotor makes use of an IMU to sense its orientation, and transmit the measurements to the estimator. Additionally, a UWB anchor, affixed to the quadrotor's body, receives ranging signals transmitted by tags, which are then processed to calculate distances. These distances are utilised to approximate the position of the quadrotor and relayed to the estimator. Meanwhile, images captured by a camera are fed into the simulator's integrated object detection algorithm, which subsequently computes an estimation of the quadrotor's position. This position data is also conveyed to the estimator. The camera and UWB-based localisation system were configured to minimise performance discrepancies along the x and y axes.

\subsection{Experimental results} \label{subsec:results}
To assess the effectiveness of the filters, we utilise the root mean square error (RMSE) of the estimated states as our performance metric. This value is calculated by comparing the estimated and true positions of the quadrotor in the x-y plane. We compute the mean, median, 75th percentile, and 25th percentile of the RMSE for each filter and scenario. To account for the stochastic nature of the system, we conduct each experiment 20 times for every filter and scenario. During these experiments, the quadrotor follows a pre-determined trajectory, which remains constant across all tests. Additionally, the quadrotor is required to maintain a constant altitude throughout the experiment.

Tables~\ref{table:exp_valid:results} and \ref{table:exp_valid:results_intermittent} show the RMSE of the estimated position of the quadrotor for the MCC-KF and the Kalman filter for two different scenarios: ($i$) Scenario 1: Measurements are continuously available from the UWB anchor and the camera. ($ii$) Scenario 2: The measurements are only intermittently available from the UWB and the camera, on a rate of approximately one measurement every ten time steps.

In the initial scenario, the MCC-KF demonstrates superior performance compared to the Kalman filter in all RMSE percentiles. This is expected, as the sensor measurements are not Gaussian, particularly the position measurements obtained from the camera, which during the experiments were found to contain a significant number of outliers. Although the Kalman filter operates successfully in this case, the MCC-KF is able to better handle this non-Gaussianity and provide a more accurate estimate. More precisely a $31.22\%$ reduction in the mean RMSE for the x position and a $30.30\%$ reduction for the y position over the Kalman filter. The second scenario provides an additional challenge to the Kalman filter, as it not originally designed to handle intermittent measurements. Despite the current implementation of the filter, which follows the approach in~\cite{hadjiloizou2022onboard} to better handle the inconsistency of the sensors, it still falls short of the MCC-KF's performance. Specifically, the latter filter reduces the mean RMSE for x position by 69.59\% and for y position by 71.76\% compared to the Kalman filter. Overall, the MCC-KF shows a similar level of improvement in performance for both x and y axes in both scenarios. Furthermore, the increase in performance improvement in the latter scenario is expected and further validates the effectiveness of the MCC-KF in handling inconsistent measurements.

After comparing the two intermittent handling measurement approaches, MCC-KF and MCC-KF-2 for the first and second handling methods respectively, it is evident that both approaches outperform the Kalman Filter. However, the former approach of MCC-KF performs better overall than the latter.

\begin{figure}[h]
    \centering
    \begin{tikzpicture}
\begin{axis}
[height=5cm,
width=9cm,
label style={font=\small},
tick label style={font=\small},
legend style={font=\footnotesize},
legend columns={-1},
legend cell align={left}, 
smooth,
point meta=explicit,
ylabel shift = 0pt,
minor tick num=1,
ylabel style={yshift=-0.4cm},
axis equal=false,
line width=0.7pt,
no markers,
grid=major,
x label style={at={(axis description cs:0.5,0)}},
y label style={at={(axis description cs:0,0.5)}},
axis on top,
xmin=-1,
xmax=14,
ymin=-0.5,
ymax=3.5,
xlabel=position - $x$~(m),ylabel=position - $y$~(m)]

\addplot+[dashed, black] table [x=xref, y=yref, col sep=comma] {results/intermittent.csv};
\addplot+[qual4] table [x=kf-x, y=kf-y, col sep=comma] {results/intermittent.csv};
\addplot+[qual5] table [x=mcckf-x, y=mcckf-y, col sep=comma] {results/intermittent.csv};

\addlegendentry{desired}
\addlegendentry{KF}
\addlegendentry{MCC-KF}
\end{axis}
\end{tikzpicture}
    \vspace{-15pt}
    \caption{Scenario 2: Intermittent communication\vspace{-0.4cm}.}
    \label{fig:intermittent}
\end{figure}
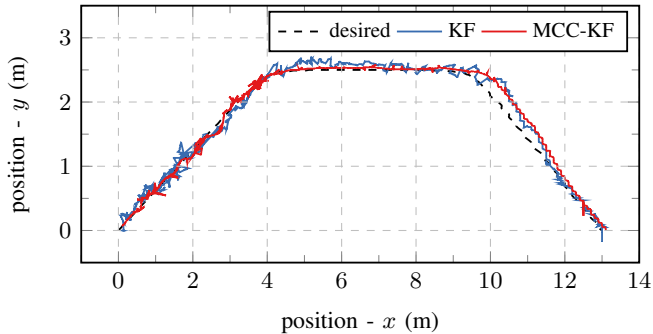

\begin{table}[h]
\begin{center}
    \caption{Estimation error with Non-intermittent Measurements ($m$)}
    \begin{tabular}{|c|c|c|c|c|}
        \hline
        & \multicolumn{2}{c|}{\textbf{Kalman Filter}} & \multicolumn{2}{c|}{\textbf{MCC-KF}} \\
        \hline
        & \textbf{x} & \textbf{y} & \textbf{x} & \textbf{y} \\
        \hline
        \hline
        \textbf{mean} & 0.0394 & 0.0297 & 0.0271 &  0.0207 \\
        \hline
        \textbf{median} & 0.0442  & 0.0297 & 0.0268 &  0.0205 \\
        \hline
        \textbf{75\%tile} & 0.0498 &  0.0301 & 0.0286 & 0.0210 \\
        \hline
        \textbf{25\%tile} & 0.0246 & 0.0291 & 0.0256 & 0.0193 \\
        \hline
    \end{tabular}
    \label{table:exp_valid:results}
\end{center}
\end{table}
\vspace{-0.7cm}
\begin{table}[h]
\begin{center}
    \caption{Estimation error with Intermittent Measurements ($m$)}
    \begin{tabular}{|c|c|c|c|c|c|c|}
        \hline
        & \multicolumn{2}{c|}{\textbf{Kalman Filter}} & \multicolumn{2}{c|}{\textbf{MCC-KF}} & \multicolumn{2}{c|}{\textbf{MCC-KF-2}} \\
        \hline
        & \textbf{x} & \textbf{y} & \textbf{x} & \textbf{y} & \textbf{x} & \textbf{y} \\
        \hline
        \hline
        \textbf{mean} & 0.0832 & 0.0602 & 0.0253 &  0.0170 & 0.0433 & 0.0303\\
        \hline
        \textbf{median} & 0.0804  & 0.0618 & 0.0234 &  0.0167 & 0.0423 & 0.0293\\
        \hline
        \textbf{75\%tile} & 0.0859 &  0.0678 & 0.0303 & 0.0180 & 0.0451 & 0.0325\\
        \hline
        \textbf{25\%tile} & 0.0717 & 0.0518 & 0.0208 & 0.0164 & 0.0364 & 0.0275\\
        \hline        
    \end{tabular}
    \label{table:exp_valid:results_intermittent}
\end{center}
\end{table}
A visual representation of an excerpt from the aforementioned results is given in Fig.~\ref{fig:non-intermittent} and Fig.~\ref{fig:intermittent} for scenario 1 and 2, respectively. In both scenarios, the MCC-KF outperforms the Kalman filter, being far less noisy overall. In the case of intermittent communication, as depicted in Figure~\ref{fig:intermittent}, the MCC-KF demonstrates its ability to track the ground truth more robustly under inconsistent measurements.

\begin{figure}[h]
    \centering
    \vspace{0.2cm}
    \begin{tikzpicture}
\begin{axis}
[height=5cm,
width=9cm,
label style={font=\small},
tick label style={font=\small},
legend style={font=\footnotesize},
legend columns={-1},
legend cell align={left}, 
smooth,
point meta=explicit,
ylabel shift = 0pt,
minor tick num=1,
ylabel style={yshift=-0.4cm},
axis equal=false,
line width=0.7pt,
no markers,
grid=major,
x label style={at={(axis description cs:0.5,0)}},
y label style={at={(axis description cs:0,0.5)}},
axis on top,
xmin=-1,
xmax=14,
ymin=-0.5,
ymax=3.5,
xlabel=position - $x$~(m),ylabel=position - $y$~(m)]

\addplot+[dashed, black] table [x=xref, y=yref, col sep=comma] {results/non-intermittent.csv};
\addplot+[qual4] table [x=kf-x, y=kf-y, col sep=comma] {results/non-intermittent.csv};
\addplot+[qual5] table [x=mcckf-x, y=mcckf-y, col sep=comma] {results/non-intermittent.csv};

\addlegendentry{desired}
\addlegendentry{KF}
\addlegendentry{MCC-KF}
\end{axis}
\end{tikzpicture}
    \vspace{-15pt}
    \caption{Scenario 1: Non-intermittent communication\vspace{-0.4cm}.}
    \label{fig:non-intermittent}
\end{figure}
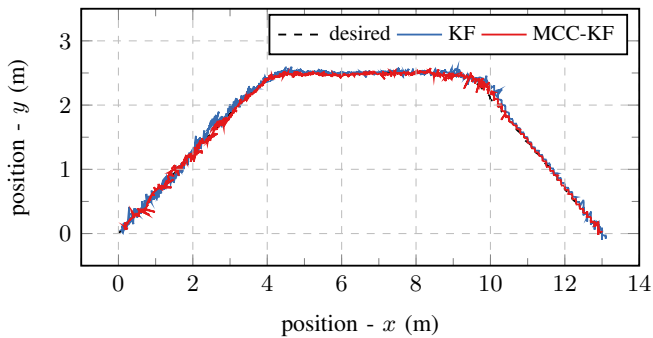

\section{CONCLUSIONS AND FUTURE DIRECTIONS}
This paper presents a multisensor fusion framework that utilises an IMU, a camera-based object detection algorithm, and an UWB localisation system for onboard navigation of a quadrotor in an indoor environment. The framework uses a MCC-KF to accurately estimate the pose of the quadrotor for tracking a predefined trajectory, albeit the non-Gaussian measurement noise and the intermittent sensor readings. 
The comparison with a previously designed Kalman filter by the authors demonstrates the effectiveness of the MCC-KF in handling indoor navigation missions. This paper 
presents a promising approach to be developed on actual quadrotor platforms towards improving the real-time and robust navigation of quadrotors in realistic complex indoor environments.

Part of ongoing research focuses on studying the stability conditions of the MCC-KF in the presence of packet drops, in the same sense as in \cite{sinopoli2004} for the classical KF. Also, considering adaptive and more advanced filtering~\cite{Qi_2023} is also under consideration.

\bibliographystyle{IEEEtran}
\bibliography{IEEEabrv,sources.bib}

\end{document}